\documentclass[10pt,twocolumn,letterpaper]{article}

\usepackage{cvpr}
\usepackage{times}
\usepackage{epsfig}
\usepackage{graphicx}
\usepackage{amsmath}
\usepackage{amssymb}
\usepackage{amsthm}
\usepackage{caption}
\usepackage{textcomp}
\usepackage{paralist}
\usepackage[linesnumbered,ruled]{algorithm2e}
\usepackage{array,multirow,graphicx}
\usepackage{setspace}
\usepackage[export]{adjustbox}

\usepackage[pagebackref=true,breaklinks=true,letterpaper=true,colorlinks,bookmarks=false]{hyperref}

\DeclareMathOperator*{\argmax}{arg\,max}
\DeclareMathOperator*{\argmin}{arg\,min}
\DeclareMathOperator*{\dom}{dom}
\newtheorem{lemma}{Lemma}

\cvprfinalcopy % *** Uncomment this line for the final submission

\def\cvprPaperID{1315} % *** Enter the CVPR Paper ID here

\ifcvprfinal\fi
\begin{document}

%%%%%%%%% TITLE
%\title{Active Learning for Multi-Label Recovery on Graphs via Adaptive Submodularity and Harmonic Analysis}
%\title{Online Graph Completion with Multi-response Measurements: Algorithm and Applications}
%\title{Online Graph Completion for Multi-response Signal Recovery via Sparse Coding}  %Multi-response Measurements: Algorithm and Applications}
\title{Online Graph Completion: Multivariate Signal Recovery in Computer Vision}

\author{
\begin{minipage}{\linewidth}
    \vspace{-15pt}
    \begin{center}
      Won Hwa Kim$^{\dagger}$, Mona Jalal$^{\mathsection}$, Seongjae Hwang$^{\dagger}$, Sterling C. Johnson$^{\ddagger}$, Vikas Singh$^{\mathsection, \dagger}$\\
%      Won Hwa Kim$^{\dagger 1}$~~~~ Hyunwoo J. Kim$^{\dagger 1}$~~~~ Nagesh Adluru$^{\ddagger}$~~~~ Vikas Singh$^{\mathsection \dagger}$\\
      \vspace{7pt}
      \textnormal {\normalsize $^{\dagger}$Dept. of Computer Sciences, University of Wisconsin, Madison, WI, U.S.A.\\
        $^{\mathsection}$Dept. of Biostatistics \& Med. Informatics, University of Wisconsin, Madison, WI, U.S.A.\\
        $^{\ddagger}$GRECC, William S. Middleton VA Hospital, Madison, WI, U.S.A.\\
%        \vspace{10pt}
        \texttt{http://pages.cs.wisc.edu/$\sim$wonhwa}
        %\texttt{ravi5@wisc.edu}~~~~\\
        %\texttt{\{scj, ozioma\}@medicine.wisc.edu}~~~~
        %\texttt{vsingh@biostat.wisc.edu}\\
%        \texttt{http://pages.cs.wisc.edu/$\sim$wonhwa}
      }
    \end{center}
    \vspace{-10pt}
  \end{minipage}
%% First Author\\
%% Institution1\\
%% Institution1 address\\
%% {\tt\small firstauthor@i1.org}
%% % For a paper whose authors are all at the same institution,
%% % omit the following lines up until the closing ``}''.
%% % Additional authors and addresses can be added with ``\and'',
%% % just like the second author.
%% % To save space, use either the email address or home page, not both
%% \and
%% Second Author\\
%% Institution2\\
%% First line of institution2 address\\
%% {\tt\small secondauthor@i2.org}
}

\maketitle
%\thispagestyle{empty}

%%%%%%%%% ABSTRACT
\begin{abstract}
%abst...
\vspace{-10pt}

The adoption of ``human-in-the-loop'' paradigms in computer vision and machine learning
is leading to various applications where the actual data acquisition (e.g., human supervision)
and the underlying inference %engine (or algorithm)
algorithms are closely interwined.
While classical work in active learning provides
effective solutions when the learning module involves classification and regression tasks,
many practical %experimental
issues such as partially observed measurements, financial constraints and even
additional distributional or structural aspects of the data
typically fall outside the scope of this treatment.
%For instance, if we perform %{\em adaptive}
%adaptive or sequential acquisition of measurements
%when the data manifest as a matrix (or tensor),
%novel strategies for completion (or collaborative filtering) have only been studied recently.
For instance, with sequential acquisition of partial
measurements of data that manifest as a matrix (or tensor),
novel strategies for completion (or collaborative filtering) of the remaining entries 
have only been studied recently.
Motivated by vision problems where we seek to annotate
a large dataset of images via a crowdsourced platform
or alternatively, complement results from a state-of-the-art object detector using human feedback,
we study the ``completion'' problem defined on graphs,
where requests for additional measurements %(supervision)
must be made sequentially.
We design the optimization model in the Fourier domain of the graph % offering substantial benefits and 
describing how ideas based on %the so-called 
adaptive %(or sequential)
submodularity provide algorithms that work well in practice.
On a large set of images collected from Imgur, we see promising
results on images that are otherwise difficult to categorize.
We also show applications to an experimental design problem in neuroimaging.

\end{abstract}

%%%%%%%%% BODY TEXT
\vspace{-10pt}
\section{Introduction}
%intro...
\vspace{-5pt}

The problem of missing or partially observed data is ubiquitous in science --- an issue that is becoming more relevant
within the translational/operational aspects of modern computer vision and machine learning.
%In typical scientific studies,
%data are first collected for various ambitious purposes.
Occasionally, we may be restricted by the number of distinct types of measurements (feedback or supervision) that can
be acquired per participant due to budget constraints.  %(e.g., on Mechanical Turk).
In other situations,
a subset (or even a majority) of features/responses may be missing in a portion of the data due to logistic
reasons. Separately, equipment malfunction, human negligence or fatigue, noise and other factors
common in data acquisition lead to scenarios where a subset of the data to be analyzed is missing,
partially observed or systematically corrupted.
%This phenomena may either be {\em prospective} or {\em retrospective}.
Occasionally, this phenomena may be {\em prospective} --- a design choice where the experiment can acquire extensive supervision only for a few samples.
%indeed be a design choice where the experiment requires extensive supervision only for a few samples.
%(e.g., the experiment can acquire extensive supervision only for a small number of samples).
Alternatively, it may be a nuisance that must be accounted for in a {\em retrospective} manner (e.g., $10\%$ of participants
labeled merely half of the objects in the image).
As the number of computer vision and machine learning systems deployed in the real world %scenarios
continues to grow and ``human-in-the-loop'' paradigms %in vision and machine learning
become mainstream,
such issues will emerge as a first order constraint that should inform the design of algorithms.
%logistic problems

%In most data acquisition studies, budget constraints, equipment malfunction, human negligence, noise and numerous other factors
%lead to
%However in practice, there are many obstacles in the data acquisition process (e.g., budget constraint, equipment malfunction, human introduced error and noise)
%which yield either missing or meaningless measurements.
%therefore many datasets end up with missing measurements.
%This becomes a serious problem in the experimental design phase,
%where many of the samples with partially missing data are often discarded or synthetic data are imputed to fill in the blanks \cite{efron1994missing,rubin1976inference}.
%When these missing data are not carefully handled especially without knowing the source of the missingness,
%its downstream analyses will introduce bias or strongly affect its statistical outcome \cite{allison2001missing}.
%In any case, without the ability to fully utilize the data,
%the follow-up analysis eventually ends up with sub-optimal results with significantly decreased statistical power.
%not being able to full utilization of the data
%% In many cases, Even after collecting for ambitious objectives,
%% a substantial
%% Depending on how the data are missing (e.g., missing at random and missing based on unobserved predictor), traditional approaches impute the missing data or perform regression to fill in the missing data.

{\em Example 1.} We are tasked with collecting human annotations on 1M+ images, via a
crowdsourced platform. The allocated budget, unfortunately, is only enough for 500K image-wise annotations.
%Budget only for 500K.
Assume that 250K randomly selected images in the corpus have already been annotated in the first phase.
%An interesting question we may ask is: % the following.
We may ask an interesting question:
based on image features calculated
(e.g., using a deep network \cite{tang2011image,yang2016improving,kim2016semi}) on the full dataset,
if we could only acquire partial data
based on financial constraints,
can we come up with a ``policy'' to decide which subset of 250K images should we request user feedback on?
Is one `order' of requests (policy)
better than the other?
If we know that we will run a simple logistic regression using the annotations
%(possibly unrelated to the image features derived for a categorization task),
what properties of the data will ensure that we obtain guarantees on the downstream machine learning model?
%In other words, the model learned from our specific choice of 500K is no more than $\epsilon$ different from the true one.

{\em Example 2.} Consider the setting where we have access to an (already trained) model for object detection.
When we use this system on images obtained via a platform such as Reddit or Imgur,
it works well but fails for $\eta\%$ of the images.
%Object detectors on images, e,g., on Reddit or Imgur. Works well but fails. that's fine.
Let us assume that the already learned system offers good specificity,
i.e., when the model is highly confident, its predictions correlate with ground truth labels.
Separately, we also have auxiliary information (e.g., comments, captions associated with each image).
While not perfect, such secondary data provide some sense of associations between images.
%Given that this side information is informative (although not perfect),
If this were a partially observed distribution (with $\eta\%$ of missing observations), can we provide new object probabilities on
images where a state of the art object detectors failed?
Now, if human supervision were available to annotate a small portion, $\eta\%$ of images,
in which order of images will we ask the human to intervene? 
Thinking of object-wise probabilities as a multivariate ``signal'', can the signal on the remaining subset be ``completed''?

{\em Example 3.} In a neuroimaging study, we may be provided a set of relatively cheaper measurements (e.g., MRI scans) on all subjects in a cohort.
Let us assume that these measurements are correlated with a more expensive and highly informative acquisition such as a PET scan;
summaries obtained from the less expensive scans are useful but have higher variability \cite{weigand2011transforming}.
%but the cheaper measures
%have far more variability.
What can be gleaned from the data statistics of the cheaper set of imaging measures?
How can such information guide the sequential order
in which more expensive data will be collected on the remaining participants with budget constraints?
Can we guarantee that the statistical power of the downstream model will improve? % (over a random selection of participants)?

If we ignore the online %(or sequential policy)
aspect of the problems above, it is reasonable to think of examples 1 to 3 above via
the lens of matrix completion \cite{donoho2006compressed,candes2010matrix,candes2009exact,yu2009fast,candes2009exact,cai2010singular,ji2010robust}.
%To deal with the issues of missing data, matrix completion methods are broadly deployed in computer vision and machine learning.
%to deal with the issues of missing data.
Indeed, each subject/participant can be given as a column in a matrix which is partially observed (potentially corrupted) and the task
is to ``complete'' the matrix --- often, using a low rank regularizer (or its variants).
However, we %quickly
see that even the entry-level assumptions used in low rank matrix completion are violated, %above,
for instance, the restricted isometry property (RIP) and incoherent sampling.
%the de facto ways of matrix completion have led to resurgence in analysis of factorization forms ,
%which on its own has a long history since early 1990s \cite{tomasi1992shape}.
Shoehorning matrix completion schemes directly to the problem yields unsatisfactory results, as we will describe shortly.

\begin{figure}[!tb]
%  \vspace{-18pt}
  \centering
%  \scalebox{0.7}{
%  \includegraphics[width = .2\textwidth, height = 0.16\textheight]{fig/arma_mesh2}
  \includegraphics[width = .31\columnwidth, height = 0.11\textheight]{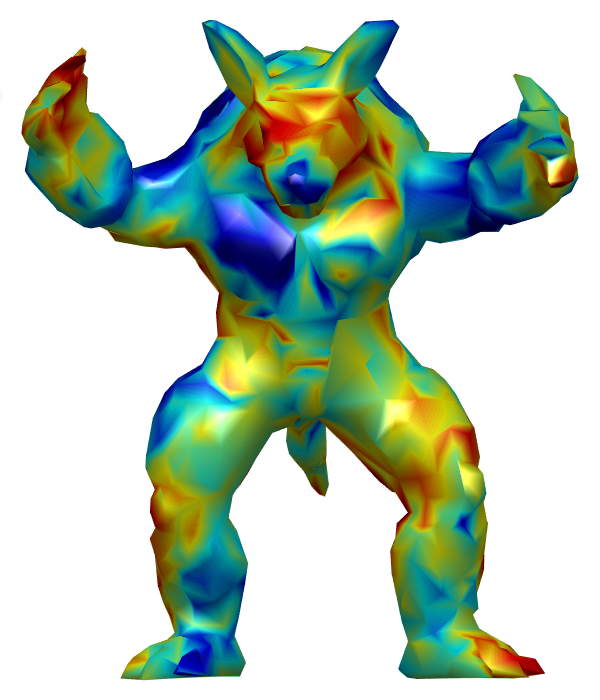}\hspace{-3mm}
  \hspace{1mm}
  \includegraphics[width = .31\columnwidth, height = 0.11\textheight]{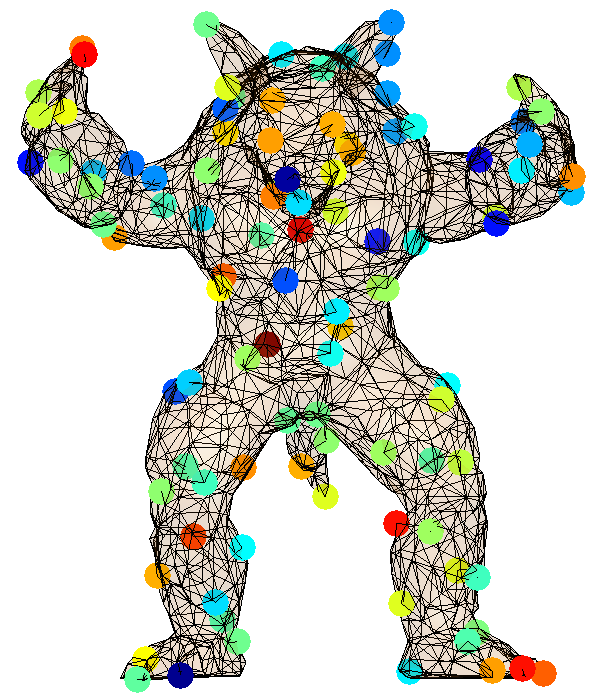}\hspace{-3mm}
  \hspace{1mm}
  \includegraphics[width = .31\columnwidth, height = 0.11\textheight]{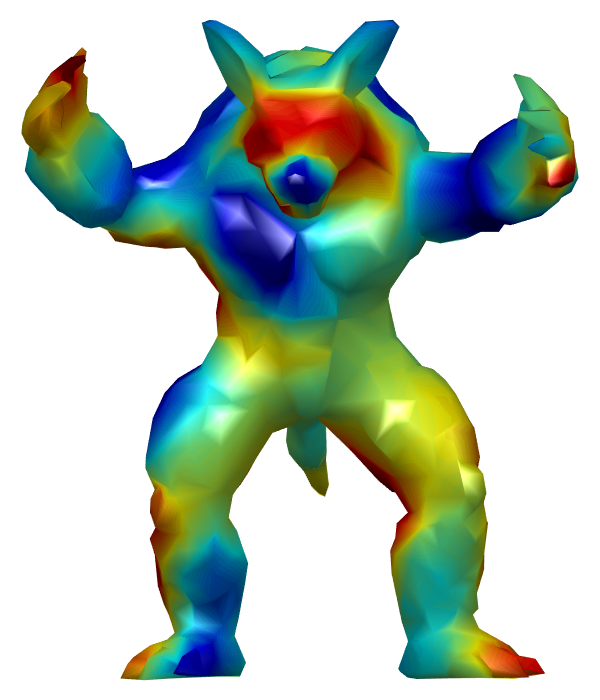}%\hspace{-3mm}
  \includegraphics[width = .05\columnwidth, height = 0.1\textheight]{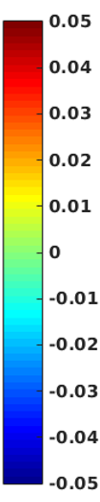}
%  }
  \vspace{-10pt}
  \caption{\footnotesize An example of graph completion on Armadillo mesh, given edge weights based on curvature.
    Left: noisy signal on the mesh,
    Middle: partial observation on the signal,
    Right: recovery of the signal on the mesh.
%From left to right: a graph, a signal (measurement) defined on the graph, partial observation on the signal and recovery of the signal based on the partial measurement.
}
  \label{fig:arma}
  \vspace{-20pt}
\end{figure}

{\em Graph representation.} Instead of a matrix, it is perhaps more natural to express the data in terms of a graph.
Individual participants are nodes and their measurements can be assumed to be an observed
multivariate signal of dimension $p$ on each node.
If we assume some auxiliary information yields associations between these nodes,
then the partially observed setting models the situation that at some nodes
we do not observe the signal at all%(or only observe a small subset of $p$ measures)
, see Fig. \ref{fig:arma}.
%Now, consider a variation of this problem where the structure is slightly harder to capture in terms of parsimony.
%That is, a matrix (or an image) has a nice regular structure to begin with,
%but the question is whether we can work on a similar task without the regularity in the domain.
%Such a complex domain is generally represented
%as a graph with vertices representing data points and edges that are connecting them,
%and measurements are defined as signals on the vertices.
%When $N$ number of graph vertices are enumerated and corresponding multi-variate measurements of size $p$ are laid next to each other,
%they compile a form of matrix of size $N \times p$.
%With missing measurements in this matrix, it clearly yields a matrix completion problem.
%However in this case, the data are missing at certain rows instead of random locations and the relationships between elements no longer hold as in traditional matrix completion.
%In this regime, the aforementioned assumptions become ambiguous and it becomes difficult to achieve a feasible solution for
%this {\em graph completion problem}.

This ``discrete'' space (i.e., graph) version of completion problems has only been studied/formalized within the last two years.
In \cite{rao2015collaborative},
the idea of collaborative filtering was generalized to graphs where a smoothness assumption was imposed using the Laplacian of the graph.
Separately, a random sampling scheme with a bandlimited assumption was introduced in \cite{puy2015random} where the authors define a probability distribution for sampling
at each node of a graph by analyzing the eigenvalues/eigenvectors of the Laplacian of the graph.
These methods essentially model the graph completion problem (an example demonstrated in Fig.~\ref{fig:arma}) using a diffusion process %which is a smoothing operation
by propagating observed measurements to their neighboring vertices where the measurements are unobserved.
%% A similar operation was also acknowleged for semi-supervised learning problem on graphs as well,
%% where known labels are propagated to the vertices with unknown labels \cite{kim}.
%In either approaches,
They utilize the spectrum of the Laplacian of a graph to simulate the diffusion process
in the native space (i.e., a graph),
and solve an optimization problem {\em in the graph space} to obtain the optimal solution.
%We would like to explore the same
%Although these methods have paved the way to generalize traditional signal recovery problems to a graph domain,
%these solvers do not take advantage of the bandlimited nature of signals nor the low-rank property
%%as in other matrix completion approaches.
These are important results which provide baselines for our experiments.  %when we present our experimental results.

{\em Key Ideas.} The starting point of our proposed algorithm is to perform harmonic analysis on the given graph.
Similar to the ``low rank'' property (for matrix completion), we also make use of parsimony/sparsity, albeit in terms of
representations obtained in the Fourier/wavelet space of the graphs.
Recall that measurements/signals are represented as a smooth function in their %native
graph space but their representations in a dual space %(e.g., frequency space)
may be sparse,
%this is the intuition behind Fourier/wavelet analysis in signal processing.
which is an important advantage of the frequency analysis \cite{kim2016adaptive}.
We exploit a similar idea, in the graph setting using the graph Fourier transform.
The ``online'' version of the completion problem is defined using the frequency space of this graph.
When we acquire a measurement on a vertex, the ``value of information'' for the remaining set of unobserved vertices {\em changes}
%% and the utility/objective function is adjusted and will update the remaining graph nodes
to impact %/modify
our ``policy'' to acquire the next measurement.
This strategy is related to the idea of diminishing returns but is an ``adaptive'' variation. %, which poses interesting technical challenges.
%in terms of optimization and active learning.
While such an online scenario has been studied for a general matrix or tensor setting \cite{krishnamurthy2013low,kumar2012sampling,lin2011linearized},
no algorithms are available for graphs. % setting.
We show how recent work on submodular maximization can be adapted for analysis of measurements on a graph in this online manner
utilizing the graph Fourier representation.

In this paper, we provide a framework for deciding the optimal policy of selecting vertices on a graph
for an accurate and efficient recovery of a signal by exploring its dual representation.
The {\bf contributions} of this paper are:
1) we propose an algorithm for sequentially selecting vertices on a graph using adaptive submodularity,
2) we provide an algorithm for sequentially recovering signals on graph vertices using the graph Fourier transform,
3) we demonstrate extensive results on large-scale image datasets as well as a neuroimaging dataset.
On the image data, we estimate object labels on images where state-of-the-art object detectors fail.
On the neuroimaging data, we estimate expensive summary measures from brain scans using other cheaper measures.

\vspace{-8pt}
\section{Background: Fourier and Wavelet Transforms in Non-Euclidean Spaces}
\label{sec:prelim}
\vspace{-5pt}
%% Linear (forward/backward) transforms of signals
%% such as Fourier and wavelet transforms are well known in the Euclidean setting.
%% However, these

The Fourier and wavelet transforms have been extensively
studied almost exclusively in Euclidean spaces.
%adoptable for analyses of data in non-Euclidean spaces
Recently, several groups have demonstrated the analogs of these transforms
in non-Euclidean spaces \cite{Coifman200653,Hammond2011129}, which are fundamental for our proposed algorithm.
We therefore provide a brief description in this section.  %for both the Euclidean and non-Euclidean settings.

\vspace{-5pt}
\subsection{Fourier and Wavelet Transforms}
\vspace{-5pt}
%% The Fourier transform is a fundamental tool for frequency analyses of a signal
%% by transforming the signal $f(x)$ into the frequency domain ashist
The Fourier transform transforms a signal $f(x)$ in $x$ to $\hat f(\omega)$ in the frequency space $\omega$
using $\sin()$ basis functions as
%% by transforming the signal to the frequency space as
%% \begin{equation} %\footnotesize
%% \label{eq:fourier}
%\hat f(\omega) = \underbrace{\int f(x) \mathrm e^{-j\omega x} \mathrm d x}%_{\langle f, {\rm basis} \rangle}
$\hat f(\omega) = \langle f, e^{j\omega x} \rangle = \int f(x) \mathrm e^{-j\omega x} \mathrm d x$.
%\end{equation}
%% where $\hat f(\omega)$ is the signal in the frequency space known as Fourier coefficient.
The concept underlying the wavelet transform is similar
but %instead of the Fourier bases (i.e., $\sin()$),
it utilizes a localized oscillating basis function (i.e., mother wavelet) for the transform.
%to transform a signal to a dual domain for frequency analysis
While the Fourier basis has an infinite support,
a wavelet $\psi$ is localized in both time and frequency space \cite{mallat1999wavelet}.
%% It has two fundamental properties, {\em scale}
%% and {\em translation} $a$ properties. % in the original domain
%% %In the frequency domain, $\psi$ behaves as a localized band-pass filter,
%% Here, changes in $s$ and $a$ control the dilation and the location of $\psi$ respectively.
A mother wavelet with {\em scale} $s$ and {\em translation} $a$ parameters
is written as $\psi_{s,a}(x) = \frac{1}{s} \psi (\frac{x-a}{s})$, where
%where controlled by parameters $s$ and $a$ respectively, .
%% has {\em scale} and {\em translation} properties controlled by
%% parameters $s$ and $a$ respectively.
changing $s$ and $a$ varies the dilation and location of $\psi_{s,a}$ respectively.
%% %In the frequency domain, $\psi$ behaves as a localized band-pass filter,
%$s$ controls the dilation and varying $a$ controls the location of $\psi$.
Using $\psi_{s,a}$ as basis, a wavelet transform of a function $f(x)$ yields wavelet coefficients $W_f(s,a)$ defined as
%results in wavelet coefficients $\mathcal  W_f(s,a)$ at scale $s$ and at location $a$ as
\begin{equation} %\footnotesize
  \small
  \label{eq:wc}
  W_f(s,a) = \langle f, \psi \rangle = \frac{1}{s} \int f(x) \psi^*
  (\frac{x-a}{s}) \mathrm{d}x
\end{equation}
where $\psi^*$ is the complex conjugate of $\psi$.

%Interestingly, $\psi_s$ is localized not only in the original domain but also in the frequency domain.
In the frequency space, $\psi_s$ behave as band-pass filters covering different bandwidths corresponding to scales $s$.
%For the case where these band-pass filters do not cover the low-frequency regions,
When these band-pass filters do not handle low-frequency bands,
a scaling function $\phi$ (i.e., a low-pass filter) is introduced.
In the end, a transform of $f$ with the scaling function $\phi$ results in a smooth representation of the original signal
and filtering at multiple scales $s$ of the mother wavelet $\psi_s$ offers a multi-resolution view of the given signal.
In both cases for the Fourier and wavelet transforms,
there exist inverse transforms that reconstruct the original signal $f(x)$
using their coefficients and the basis functions. % that were used for the transforms.

\vspace{-5pt}
\subsection{Fourier and Wavelet Transforms for Graphs}
\vspace{-5pt}
%In the Euclidean space,
The Euclidean space is typically represented as a regular lattice,
therefore one can easily construct a mother wavelet with a certain shape
to define a wavelet transform.
%In this case, one can easily define the shape of a mother wavelet in the context of an application.
On the other hand, in non-Euclidean spaces that are
generally represented by a set of vertices and their arbitrary connections,
%(e.g., graphs that consists of a set of vertices and edges with arbitrary connections),
the construction of a mother wavelet is ambiguous due to
the definition of dilation and translation of $\psi_{s,a}$.
Because of these issues, %although wavelet transform has been very popular in Signal Processing and Computer Vision,
the classical %definition of
Fourier/wavelet transform has not been suitable for analyzing data in complex space until recently
when \cite{Hammond2011129,Coifman200653} proposed these %Fourier and wavelet
transforms %on graphs as well as in other
for graphs.

The core idea for constructing a mother wavelet $\psi_s$ on the nodes of a graph comes from
its representation in the frequency space.
By constructing different shapes of band-pass filters in the frequency space and
transforming them back to the original space,
we can %bypass the ambiguity of scale and translation in the original space and
implement mother wavelets that maintain the traditional properties of wavelets.
%Instead of defining its shape in the original domain where the properties of $\psi$ are ambiguous,
%we define a mother wavelet in a dual domain where its representation is clear and then transform it back to the original domain.
Such an implementation requires a set of ``orthonormal'' bases and a kernel (filter) function. % in frequency space.
The orthonormal bases span the analog of the frequency space in the non-Euclidean setting and
the kernel function denotes the representation of $\psi_s$ in the frequency space.
%2) a kernel function $g()$ to define a band-pass filtering operation to decide the shape of $\psi$.
%By defining a band-pass filter $h()$ in the frequency space and localizing with a $\delta_n$ using the orthonormal bases,
%% It is simple to design a filter function $g()$ in the 1D frequency domain,
%% and we can use spectral graph theory to achieve the orthonormal bases for a graph %\cite{Hammond2011129}
%% to define Fourier transform for a graph.
In this sense, when the non-Euclidean space is represented as a graph,
we can adopt spectral graph theory \cite{chung1997spectral} for orthonormal bases and design a $g()$ in the space spanned by the bases.

In general, a graph $\mathit G = \{ V, E\}$ is represented by a set of vertices $V$ of size $N$ %number of vertices
and a set of edges $E$ %with edges
that connects the vertices.
An adjacency matrix $\mathit A_{N \times N}$ is the most common way to represent a graph $\mathit G$
where each element $a_{ij}$ denotes the connection in $E$ between the $i$th and the $j$th vertices by a corresponding edge weight.
Another matrix, a degree matrix $\mathit D_{N \times N}$,
is a diagonal matrix
where the $i$th diagonal element is the sum of edge weights connected to the $i$th vertex.
From these two matrices, a graph Laplacian is then defined as $\mathit L = \mathit D - \mathit A$.
Note that $\mathit L$ is a self-adjoint and positive semi-definite operator, therefore
%The matrix $\mathcal L$ can be decomposed into
provides pairs of eigenvalues
$\lambda_l \geq 0$ and corresponding eigenvectors $\chi_l$, $l = 1,\cdots, N$ which are orthonormal to each other.
The bases $\chi$ can be used to define the {\em graph Fourier transform} of a function $f(n)$ defined on the vertices $n$ as
%to define the graph Fourier transform of a function $f(n)$ defined on the vertices $n$ as
\begin{equation} %\footnotesize
  \small
  \hat f (l) = %\langle \chi_l, f \rangle =
  \sum_{n=1} ^{N} f(n) \chi^* _l (n) ~~~\mbox{and}~~~
  f (n) = \sum_{l=1} ^{N} \hat f(l) \chi _l (n)
\label{eq:gft}
\end{equation}
where $\chi^*$ is a conjugate of $\chi$.
Here, the graph Fourier coefficient $\hat f(l)$ is obtained by the forward transform
and the original function $f(n)$ can be reconstructed by the inverse transform.
If a signal $f(n)$ lives in the spectrum of the first $k$ eigenvectors, % in the dual space,
we say that $f(n)$ is $k$-bandlimited.
This transform offers a way to look at a signal defined on graph vertices
in a dual space which is an analog of the frequency space in the Fourier transform. %just as the conventional Fourier transform.

\begin{figure}[!tb]
  \centering
  \includegraphics[width = 0.3\columnwidth]{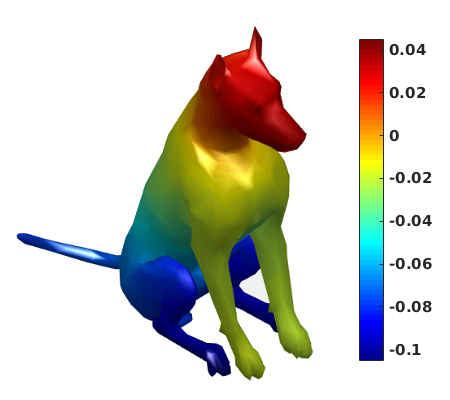}
  \includegraphics[width = 0.3\columnwidth]{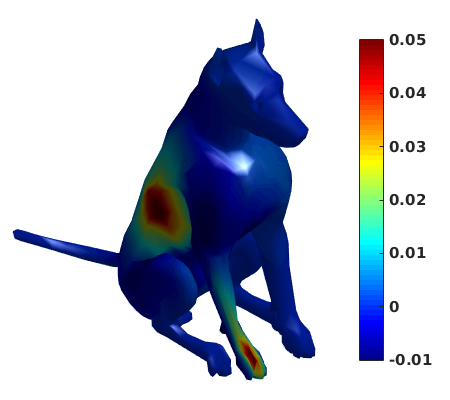}
  \includegraphics[width = 0.3\columnwidth]{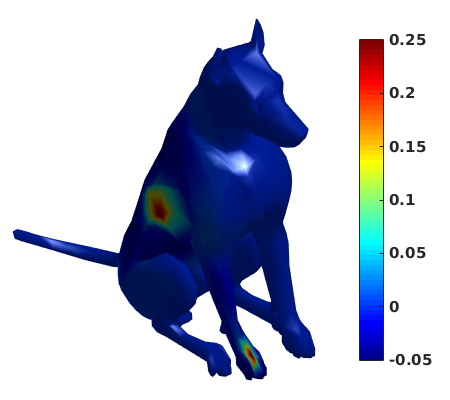}
  \vspace{-15pt}
  \caption{\footnotesize Examples of basis functions on a dog shaped mesh.
    Left: $\chi_3$ (not localized), Middle: two $\psi_1$ (localized) at the back and on a paw, Right: $\psi_5$ (localized and condensed). }
  \vspace{-15pt}
  \label{fig:bases}
\end{figure}

A mother wavelet $\psi$ then can be defined using the graph Fourier transform.
First, a kernel function $g:\mathbb R^+ \rightarrow \mathbb R^+$ (i.e., band-pass filter) is designed in the dual space,
then this operation is localized by an impulse function $\delta_n$ at vertex $n$:
\begin{equation} \footnotesize
  \psi_{s,n}(m) = \sum_{l=1} ^{N} g(s \lambda_l) \chi_l ^* (n) \chi _l (m).
  \label{eq:gwavelet}
\end{equation}
Here, the scale parameter $s$ is independent from $\chi$ and defined inside of $g()$ using
the scaling property of Fourier transform \cite{sigandsysbook}. %, and it makes $s$ independent from the bases $\chi$.
Examples of localized $\psi_{s}$ on a mesh are shown in Fig.~\ref{fig:bases} comparing with a $\chi_3$ (not localized). % and localized $\psi_s$ in different scales.
%Using the graph Fourier transform,
%a mother wavelet $\psi$ is implemented by first defining a kernel function $h()$
%% and then localizing it by a Dirac delta function $\delta_n$
%% in the original graph through the inverse graph Fourier transform.
%% Since $\langle \delta_n, \chi_l \rangle = \chi_l^*(n)$,
%% the mother wavelet $\psi_{s,n}$ at vertex $n$ at scale $s$ is defined as
%% \begin{equation} %\footnotesize
%%   \psi_{s,n}(m) = \sum_{l=0} ^{N-1} h(s \lambda_l) \chi_l ^* (n) \chi _l (m).
%%   \label{eq:gwavelet}
%% \end{equation}
%% Here, using the scaling property of Fourier transform \cite{sigandsysbook},
%% the scale $s$ can be defined as a parameter in the kernel function $h()$ independent from the bases $\chi$.
%% Representative examples of a graph Fourier basis and graph wavelet bases are shown in Fig.~\ref{fig:bases}. % on a cat shaped mesh.
%% A cat shaped graph is given in Fig.~\ref{fig:bases} a), and one of its graph Fourier basis $\chi_2$  is shown in b).
%% Also, graph wavelets at two different scales (i.e., dilation) at two different locations (ear and paw) are shown in Fig.~\ref{fig:bases} c) and d).
%% Notice that $\chi$ in Fig.~\ref{fig:bases} b) is diffused all over the graph,
%% while the wavelet bases in c) and d) are localized with finite support.
%and the shape of these bases are highly dependent on the structure of the graph.

%and the eigenvalues serve as the analogs of frequency.
The wavelet transform of a function $f(n)$ on graph vertices $n$ at scale $s$
then can be written using the bases $\psi$ defined as in \eqref{eq:gwavelet},
and it follows the conventional definition of the wavelet transform yielding
wavelet coefficients $W_f (s,n)$ at scale $s$ and location $n$ as
%an inner product between the function $f$ and the mother wavelet $\psi$ as
\begin{equation} \small
  W_f(s,n) = \langle f, \psi_{s,n} \rangle = \sum_{l=1} ^{N} g(s\lambda_l) \hat f(l) \chi _l (n).
\end{equation}
This transform offers a multi-resolution view of a signal defined on graph vertices %as in traditional wavelet transform
just like %it works for signals
the traditional wavelet transform in the Euclidean space (e.g., pixels)
by multi-scale filtering.
%Our framework, to be described shortly, will utilize the definition of the mother wavelet in \eqref{eq:gwavelet} for data sampling strategy on graphs as well as the graph Fourier transform for signal recovery.
Our method to be introduced shortly will use the graph Fourier transform and wavelets on graphs
to formalize %adaptive vertex sampling strategy and recovery of the signal on graphs. % with partial observations.
adaptive vertex selection strategy and graph completion.

%% Such a transform offers a multi-resolution view of signals on graphs.
%% {\em The most important fact relevant to our formulation is that the multi-resolution property can be easily captured by
%% a single parameter $s$ in the kernel function $g$ which controls the low-pass properties of the transform entirely.}

%{\em Remark.} The goal of our framework is to

%Such a wavelet transform on graphs, known as the recently proposed Spectral Graph Wavelet Transform (SGWT) \cite{Hammond2011129}, is not novel to this paper but has been limited to only few problems in vision and machine learning \cite{kim2015statistical,kim2013multi,kim2012wavelet}.
%to our knowledge has not been utilized much for vision and machine learning problems.
%since it uses
%the bases from spectral graph theory.

\vspace{-8pt}
\section{Our Proposed Algorithm}
\vspace{-5pt}
Consider a setting where there exists an unknown bandlimited signal $f$
of $p$ features defined on $N$ graph vertices (in an identical state space).
In other words, at each vertex $v$, we can in principle obtain a $p$-dimensional feature.
However in reality, we may be able to observe the signal only at $m$$\ll$$N$ different vertices of the graph
due to budget constraints.
In this setting, there are two core questions we may ask related to the recovery of the signal $f$ at {\em all} vertices:
1) how to efficiently recover the signal on every vertex and
2) how to select the best $m$ vertices (and in which order) to acquire the additional measurements.
We tackle these problems
%in an iterative manner
by formulating an adaptive submodular function derived from the frequency space of the graph.
%\textcolor{red}{The reader will now notice that our approach is slightly different from active learning setting where they...}
%active learning
%which do not deal with %falls outside of the scope of %many practical issues such as
%partial measurements and
%missing data nor structure of the data. %, typically fall outside the scope of this treatment.
%but closer to the adaptive matrix completion \cite{krishnamurthy2013low}.
%The reader will notice that our approach is slightly different from the setting in active learning, where they
We provide our solutions to the two questions by showing that our formulation is {\em adaptive submodular}
and proposing an algorithm %utilizing these two ideas
to recover the full signal.  % using an active learning scheme.
Notice the distinction from classical active learning (also see \cite{kim2016adaptive,krishnamurthy2013low,kumar2012sampling}) that
%which offers solutions when the learning module involves classification and regression tasks,
our specification is agnostic to the subsequent task (e.g., classification).

\vspace{-3pt}
\subsection{Signal Recovery in Graph Fourier Space}
\vspace{-5pt}
\label{sec:recovery}
Suppose we have collected data from $m$ number of vertices $Y \in \mathbb R^{m \times p}$ from a full (unknown) function $f \in \mathbb R^{N\times p}$. 
Here, our objective is to recover the original signal $f$ based on the partial observation $Y$. 
We denote the set of selected indices as $W = \{ w_1, w_2, \cdots, w_m\}$, and define a projection matrix $P$ that maps $f$ to $Y$ (i.e., $Pf = Y$):  
\begin{equation} 
\small
\label{eq:M}
  P_{i,j} = 
  \begin{cases}
    1~~~ \text{if} ~~~j = w_i \\
    0~~~ \text{o.w.}
    \end{cases}. 
\end{equation}
Based on the data $Y$ from the selected data points (i.e., vertices), 
a naive signal recovery algorithms may solve for 
\begin{equation}
\small
  Z^* = \argmin_{Z \in \mathbb R^{N\times p}} ||PZ-Y||_{\ell_2}^2 %+ \gamma g^{T} h(\mathcal L) g
  \label{eq:opti1}
\end{equation}
which minimizes the error between the observation and the estimation and 
%typically used with additional constraints (e.g., smoothness). % $z^{T} g(\mathcal L) z$). 
typically used with a smoothness constraint. 
However, such a formulation 
%may not be efficient since it 
operates in the native space of $\mathbb R^{N\times p}$  %and the full diagonalization of $L$. 
without utilizing the {\em bandlimited} property of signals. 
%without utilizing a fundamental property of signals: bandlimitedness. 
%assumption of $k$-bandlimited property of the original signal. 
%Also, it may be computationally challenging to deal with a smoothness constraint that may require full diagonalization of $\mathit L$. 
It can be also computationally challenging to deal with other constraints that requires full diagonalization of $\mathit L$. 
%It also requires an additional term in the cost function (e.g., smoothness constraint such as $g^{T} h(\mathcal L) g$) to fill in the rest of the vertices which may make the problem more computationally challenging. 
%Utilizing the assumption of $k$-band limited property, we take this problem in to its dual space
We therefore take this problem into its graph Fourier space using a set of orthonormal bases $U_k = [\chi_1, \chi_2, \cdots, \chi_k]$ 
and search for a solution in the dual space spanned by $U_k$. 
%% Transforming \eqref{eq:opti1} into the frequency space using the graph Fourier transform by $U_k$ (i.e., analogues of $\sin()$ bases), it becomes
%% \begin{equation}
%% %\footnotesize
%%   \small
%%   \hat Z^* = \argmin_{\hat Z \in \mathbb R^{k\times p}} ||PU_k\hat Z-Y||_{\ell_2}^2.  %+ \gamma g^{T} h(\mathcal L) g
%%   \label{eq:opti2}
%% \end{equation}
One of the most fundamental properties of the Fourier representation is its sparsity.  
In many cases, even a very dense form of signals in its original domain can be reconstructed with a few $\sin()$ functions.  
%This is because 
Signals in the image space tend to be smooth among pixels that are spatially close, 
on the other hand, their frequency representations are independent from such a spatial constraint \cite{kim2016adaptive}.  
Such an observation is crucial for methods for low-rank estimation of signal/measurement and has been utilized in machine learning and computer vision literature \cite{candes2009exact,cai2010singular}. 
In this regime, %since we are solving for a bandlimited solution, 
we would want to obtain a bandlimited solution that is sparse within the range of $k$-band. 
Transforming \eqref{eq:opti1} into the space spanned by $U_k$ and imposing $\ell_1$-norm constraint for the sparsity, we obtain % on our problem in the dual space, we obtain 
\begin{equation}
%\footnotesize
  \small
  \hat Z^* = \argmin_{\hat Z \in \mathbb R^{k\times p}} ||PU_k\hat Z-Y||_{\ell_2}^2  + \xi||\hat Z||_{\ell_1}
  \label{eq:opti3}
\end{equation}
where $\xi$ controls the sparsity and its solution is easily obtainable using a LASSO solver \cite{tibshirani1996regression}. 
The optimal solution $\hat Z^*$ here is an estimation of sparse encoding of the original signal $f$ 
in the frequency space, and its representation in the original space can be empirically recovered by performing 
the inverse graph Fourier transform as $Z^*(n) = \sum_{l=1}^k \hat Z_k^*(l) \chi_l(n)$. 
Note that in \eqref{eq:opti3}, we avoid imposing a smoothness constraint that has been used in other approaches \cite{rao2015collaborative,puy2015random}, 
since our solution is already smooth due to its low-rank and bandlimited properties. 
%it is expected to be smooth. 
%the smoothness constraint becomes unnecessary, 
However, the smoothness criteria may still be useful when our assumption (i.e., sparsity) in the dual domain does not hold. 
% and it will complicate the problem to become a group LASSO problem. 

\vspace{-3pt}
\subsection{Performing Adaptive Selection of Vertices}
\vspace{-5pt}
\label{sec:selection}
In order for our signal recovery process
to obtain the best estimation possible, the optimal sequential selection
of vertices to construct the projection matrix $P$ is critical.
For this task, we approach this problem from an {\em adaptive submodular} perspective.
Let us first clarify some notations to describe adaptive submodularity.

Given a set of vertices $V$ with possible states $S$,
we denote a function $\gamma: V \rightarrow S$ as a {\em realization}.
We also denote $\Gamma$ as a random realization with a prior probability $\mathbf p(\gamma) := \mathbb P [\Gamma = \gamma]$.
Under this setting,
we look for a strategy to select a vertex $v$, observe its state $\Gamma(v)$ and then select the next vertex conditioned on the previous observations.
The set of observations until the most recent stage is represented as partial realization $\theta$ and its domain defined as $\dom(\theta) = \{v | \exists o, (v,o) \in \theta \}$.
The selection process defines a policy $\pi = \{ \pi_1, \pi_2, \cdots \pi_m\}$ which is an ordered set of $m$ number of selected vertices.
Given a policy $\pi$, a function $f:2^V \times O^V \rightarrow \mathbb R$ depends on the selection of vertices and its states.
Defining $V(\pi, \Gamma)$ as the set of vertices under realization $\Gamma$,
we can formulate a problem to identify the optimal strategy
with $f_{avg} =  \mathbb E [f(V(\pi, \Gamma)), \Gamma]$ as
%\begin{equation}
%{\small
\begin{equation}
  \small
  \pi^* \in \argmax_{\pi} f_{avg}(\pi) \text{~~s.t.~~} |V(\pi, \Gamma)| \leq k \nonumber
\end{equation}%}%
%\end{equation}
that is known as {\em adaptive stochastic maximization} problem \cite{golovin2011adaptive}.
With conditional expected marginal benefit defined as
\begin{equation}
  \small
  \Delta (v|\theta) = \mathbb E [f(\dom(\theta) \cup \{v\}, \Gamma) - f(\dom(\theta),\Gamma)|\Gamma],
  \label{eq:cond_benefit}
\end{equation}
it is known that
a function $f:2^V \times S^V : \rightarrow \mathbb R$ is {\em adaptive monotone}
when $\Delta (v|\theta) \geq 0$
and {\em adaptive submodular} when
$\Delta (v|\theta) \geq \Delta (v|\theta')$ with $\theta \subseteq \theta'$
\cite{golovin2011adaptive}.
Such a problem is easily solved approximately by a greedy algorithm that maximizes $\Delta (v|\theta)$
at each iteration.
For example, if the $V$ are potential locations to place certain sensors
and $f()$ is a function that computes the area covered by the sensors,
given a probability that some sensors fail at random (i.e., $\mathbf p(\gamma)$),
one can maximize the total expected area covered by selected sensors by such an algorithm.

%% \begin{figure*}[!htb]
%%   \centering
%%   \includegraphics[width = .23\textwidth]{fig/ground_truth}
%%   \includegraphics[width = .23\textwidth]{fig/50samples}
%%   \includegraphics[width = .23\textwidth]{fig/110samples}
%%   \includegraphics[width = .23\textwidth]{fig/170samples}
%% \end{figure*}

Such a setup can be computationally challenging due to the size of $V$
%(as we will see in our experiment on a large-scale image data)
and requires an accurate prior probability.
We therefore tackle our problem of selecting the vertices in a simpler manner by computing a ``leverage value''
that describes the importance of each vertex using frequency properties of a given graph. %for our signal recovery process.
In our formulation, once a vertex is selected based on the leverage measure and data are acquired,
then its state gets fixed
(i.e., placed sensors do not fail).
Notice that such a setting makes the problem deterministic. % rather than stochastic.
However, once we observe the state of the vertex and evaluate
%how much information it can contribute
its contribution
to the signal recovery process,
we will adaptively modify the leverage value for all remaining vertices to make the next selection. %and thereby, the next selection.  %for the next selection.
That is, once a vertex is added to the policy $\pi$,
we will perform our signal recovery process as described in section \ref{sec:recovery} to evaluate how well the signal is recovered at the newly selected vertex
%and its result
which will adaptively affect our next selection. % of the next vertex.
In this setting, the conditional marginal benefit (no longer an expectation), given a policy $\pi$ is defined as
\begin{equation}
  \small
  \Delta (v|\pi) = f(\dom(\pi) \cup \{v\}|\pi) - f(\dom(\pi)|\pi)
\end{equation}
which is a specific case of \eqref{eq:cond_benefit} with a fixed policy $\pi$ instead of a random realization.

Next, in order to define our utility function, we define a measure
that describes a notion of importance at each vertex.
At each vertex, we can define the leverage value as
  \begin{equation}
    \small
  %p(n) = ||U_k \delta_n||_2^2 = \sum_{l=1}^k \chi_l(n)^2
  I(n) = \sum_{l=1}^k g(\lambda_l) \chi_l(n)^2
  \label{eq:importance}
\end{equation}
which is a reconstruction of $\delta_n$ at vertex $n$ using $U_k$ and a kernel function $g()$ \cite{rustamov2007laplace,bronstein2010scale,aubry2011wave,hustable,kim2013multi}.
The leverage value $I(n) \geq 0$ describes how much energy is preserved at vertex $n$ at scale $s$ and
roughly describes how well a signal can be recovered at each vertex with limited number of bases.
%From a belief propagation / semi-supervised learning perspective,
In order for an accurate signal recovery on the selected vertices,
we want to prioritize vertices with high $I$ when selecting a vertex $v$ for $\pi$.
%Notice that $I$ is defined by a diffusion process defined by a kernel function $g()$,
%which means
Moreover, we assume that the signals on neighboring vertices may be similar (i.e., smooth) and
modulate down $I$s from neighboring vertices of $v$ when it gets selected.
%since information from $v$ is passed onto its neighbors.
To define the notion of ``closeness'' between vertices,
we use a diffusion-type distance \cite{coifman2006diffusion,kim2015statistical} defined as
\begin{equation}
  \small
  %p(n) = ||U_k \delta_n||_2^2 = \sum_{l=1}^k \chi_l(n)^2
  %D(m,n) = \sum_{l=1}^k e^{(-\alpha\lambda_l)} \chi_l(m) \chi_l(n).
  D_{s,n'}(n) = \sum_{l=1}^k g(s\lambda_l) \chi_l(n') \chi_l(n)
\end{equation}
which measures how far a vertex $n$ and $n'$ are at scale $s$ by an energy propagation process.
Using these concepts, given $I^j$ after $j$ number of selections,
the leverage value $I^{j+1}$ for the next selection is defined as
\begin{equation}
  \small
  I^{j+1} = I^j - \eta_j D
\end{equation}
where $\eta_j$ is a constant to set $I^{j+1}(\pi_j) = 0$.
%Notice that with $D>0$,
Notice that for the leverage values $I^j(v)$ and $I^{j+1}(v)$ on the same vertex $v$ at $j$ and $(j+1)$th iterations, $I^j(v) \geq I^{j+1}(v)$ with $D>0$.
With the leverage value $I^j$ in hand,
we define a utility function $f(\pi) = \sum_{i=1}^{|\pi|} I^i(\pi_i)$
which is the sum of $I(\cdot)$ from each selection.
Using the two results below, we show that our utility function is
adaptive monotone and adaptive submodular
and can be approximately solved in a greedy way.

\begin{lemma}
\vspace{-5pt}
Given current policy $\pi = \{\pi_1, \pi_2, \cdots, \pi_j \}$ of size $j$, $f(\pi) = \sum_{i=1}^j I^i(\pi_i)$ is adaptive monotone.
\vspace{-5pt}
\begin{proof}
The conditional benefit of adding a vertex $v$ having observed $\pi$ is
{\footnotesize \begin{align}
  \Delta(v|\pi) &= f(\dom(\pi) \cup \{v\}|\pi) - f(\dom(\pi)|\pi)\nonumber\\
  &= \sum_{i=1}^j I^i(\pi_i) + I^{j+1}(v) - \sum_{i=1}^j  I^i(\pi_i)\nonumber\\
  &= I^{j+1}(v) \geq 0 \nonumber
\end{align}}
\end{proof}
\vspace{-15pt}
\end{lemma}
\noindent This lemma %is natural since
shows that the benefit of adding a vertex $v$ is always non-negative
and $f(\pi)$ follows the traditional definition of monotonicity (i.e., $f(A) \leq f(B)$ holds whenever $A \subseteq B$) with positive $I$.

\begin{lemma}
\vspace{-5pt} Given two policies $\pi$ of size $j$ and $\pi'$ of size
$j'$ where $\pi \subseteq \pi'$, our utility function $f(\pi) =
\sum_{i=1}^j I^i(\pi_i)$ is adaptive submodular.
\vspace{-5pt}
\begin{proof}
%Conditional benefit of adding a vertex $v$ having observed $\pi$ with a utility function $H()$ is
The difference between the conditional benefits from the two observations (i.e., policies) $\pi$ and $\pi'$ is
{\footnotesize \begin{align}
  \Delta(v|\pi) - \Delta(v|\pi') %\nonumber\\
  &= f(\dom(\pi) \cup \{v\}|\pi) - f(\dom(\pi)|\pi)\nonumber\\
  &~~ - (f(\dom(\pi') \cup \{v\}|\pi') - f(\dom(\pi')|\pi'))\nonumber\\
  &= I^{j+1}(v) - I^{j'+1}(v) \geq 0 \nonumber
\end{align}}%
\end{proof}
\vspace{-10pt}
\end{lemma}
\noindent This result shows that the utility function $f(\pi)$ satisfies
an adaptive analog of the traditional definition of submodularity  \cite{fujishige2005submodular}
(i.e., $f(A \cup \{v\}) - f(A) \geq f(B \cup \{v\}) - f(B)$ when $A \subseteq B \subseteq V \text{~~and~~} v \in V \setminus B$)
and it can be used to formulate an adaptive submodular optimization problem.

Given our adaptive submodular utility function at hand,
we define this iterative process as {\em Select and Recover (SR)} method by formulating the following problem:
%obtain the optimal $\pi^*$ as
{  \small
\begin{align}
  \pi^* \in &~\argmax_{\pi} ~~ \sum_{i=1}^j I^i(\pi_i)\\  &s.t.~~ I^{i+1} = I^i - \eta_i D,~ j \leq m. \nonumber
\end{align}}%
Such an adaptive submodular problem is solved by a greedy algorithm
that comes with performance guarantees \cite{golovin2011adaptive}.

Once we obtain the the optimal $\pi^*$,
we can finalize a set of selected vertices $W$ and a projection matrix $P$
which are the key ingredients for our signal recovery step. %to obtain a recovered signal over the whole graph vertices.
Using $W$, we go through the process as described in section \ref{sec:recovery} and obtain the estimation of the unknown signal. % in the frequency space of the given graph.
This whole pipeline is summarized in the Algorithm \ref{alg:alg1} below,
where we solve LASSO at each iteration which is easily scalable.
%The algorithm solves LASSO at each iteration which is easily scalable and
%% The computationally challenging component is the diagonalization of the $L$;
%% in our case, we only need to estimate $k$ eigenvectors and eigenvalues.

%\setlength{\intextsep}{0pt}
%\begin{minipage}{1.0\textwidth}
%\begin{spacing}{0.8}
%{\fontsize{8}{8}\selectfont
%  \vspace{-10pt}
  \vspace{-10pt}
\begin{algorithm}[!htb]
  \small
    \SetKwInOut{Input}{Input}
    \SetKwInOut{Output}{Output}

%    \underline{function Euclid} $(a,b)$\;
    \Input{vertex set $V$, orthonormal bases $U_k$, total number of selection $m$ and $D$ update parameter $\alpha$}
    \Output{$Z$: recovered signal}
    $\pi \leftarrow \emptyset$, $s \leftarrow 0$\\
    Derive $I^1(n)$ using $U_k$ as in \eqref{eq:importance}\\
    \For {$i=1 \text{\bf~~to~~} m$}{
    Selection step: $v^* = \argmax_v \Delta(v|\pi)$ \\ %as in section \ref{sec:selection}\\
    $\pi \leftarrow \pi \cup \{v^*\} $\\
    Observe $f(\dom(\pi))$\\
    Recovery step: obtain $Z^*$ as in section \ref{sec:recovery} \\
    $s \leftarrow \alpha|f(\pi_i) - Z^*(\pi_i)|$\\
    $I^{i+1} \leftarrow I^i - \eta_iD_s$
    }
    Return $Z^*$
    \caption{Select and Recover (SR) Method %adaptive greedy algorithm for selecting vertices and recovering signals
      \label{alg:alg1}}
\end{algorithm}
\vspace{-10pt}

\vspace{-8pt}
\section{Experimental Results}
\begin{figure}[!htb]
  \vspace{-10pt}
  \centering
  \includegraphics[trim={2cm 12.9cm 4cm 2cm},clip, width=\columnwidth]{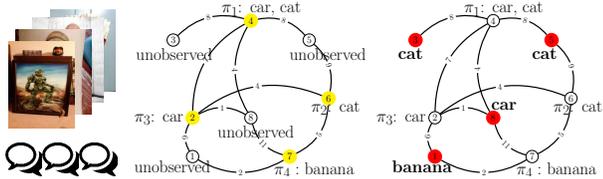}
  \vspace{-18pt}
  \caption{\footnotesize Our workflow on image datasets.
    Left: images and text, Middle: a graph derived from the text and a policy $\pi$ (yellow vertices), Right: recovered object labels on unobserved vertices (red vertices). }
  \label{fig:process}
  \vspace{-15pt}
\end{figure}

\vspace{-5pt}
In this section, we demonstrate various experimental results using our framework on three different datasets.
The first unique dataset consists of images and comments that we collected from Imgur (\url{http://www.imgur.com}),
where we qualitatively evaluate our framework for labeling objects in images where object detectors failed.
The second dataset is publicly available MSCOCO, where we make quantitative evaluations for a multi-label learning problem with human-specified object labels.  %ground truth given for object labels.
The basic schematic of how the SR method works on these datasets is shown in Fig.~\ref{fig:process}.
%The last one is a preclinical
The last experiment focuses on Alzheimer's disease (AD) image dataset that consist of participants with Pittsburgh compound B positron emission tomography (PIB-PET)
scans and Cerebrospinal fluid (CSF) data.
Here, we use the CSF measures and SR method to predict PIB imaging measures where CSF data is much cheaper to acquire.
%In all three experiments,
%% \begin{itemize}
%% \item We demostrate three different expreimental results.
%% \item Imgur dataset (we crawled), MSCOCO dataset (public) and a neuroimaging dataset.
%% \item We make qualitative evaluation on the crawled dataset, and show quantitative evaluations on the MSCOCO and neuroimaging datasets.
%% \end{itemize}

\subsection{Object Label Estimation over Object Detection}
\vspace{-5pt}
\begin{figure*}[!htb]
  \centering
  \includegraphics[trim={0 0 .6cm 0},clip, width = 0.83\textwidth]{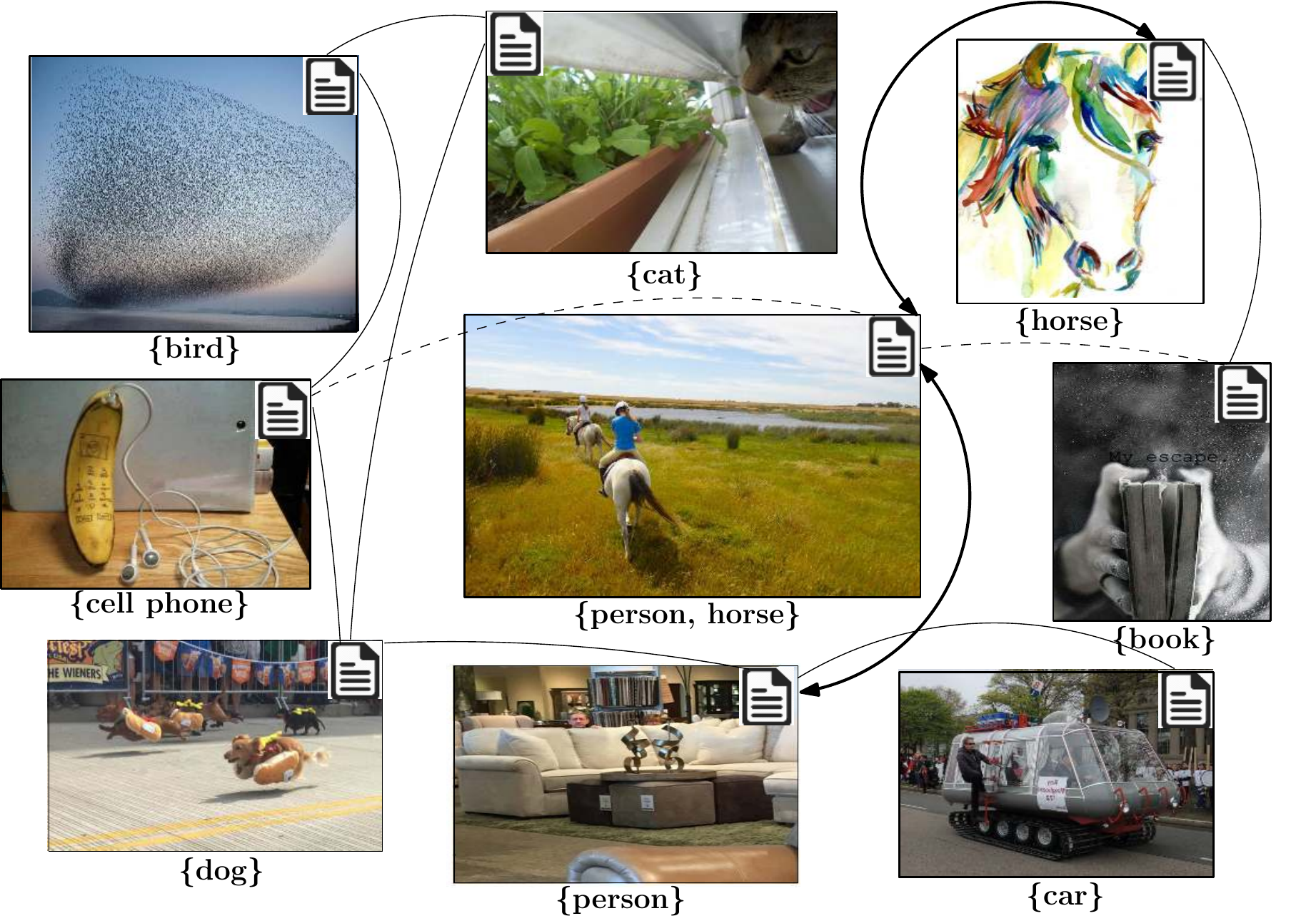}
  \vspace{-10pt}
  \caption{\footnotesize Various results on object label estimation from our Imgur experiment.
      YOLO did not confidently assign any labels on these images (i.e., below 40\% confidence) using our 75 categories.
      However, our framework suggested that there were some objects in these image. The images represent nodes and the lines denote edges in our framework,
and there are strong relationships between images with same object labels. }
  \label{fig:result_imgur}
  \vspace{-18pt}
\end{figure*}

\label{sec:imgur}
{\bf Dataset.}
%We first crawled images and comments from Imgur to generate a unique dataset to test our framework.
%To download images and comments we used Imgur REST API as well as imgurpython Python library to retrieve posts that include a single jpeg image.
%For consistency with our MS-COCO experiment,
We used MSCOCO categories to select a subset of categories on Imgur which provided images and comments, which gave us an
interesting dataset to evaluate our algorithm.
%% We downloaded the Imgur images and comments from the categories that overlaps with MSCOCO categories
%% to generate a unique dataset to test our framework. % using Imgur REST API as well as imgurpython Python library.
For each image, we obtained the top 10 comments upvoted by the community.
%and neglected the posts with less than 10 comments.
We also created a dictionary of most commonly used words on Imgur
(e.g., upvoted and downvoted)
which were
%and
removed from the comments. % for simplicity. % and increasing accuracy of our algorithm.
%Considering the fact that our target number of downloaded images was 10k and Imgur API has a rate limit of 12500 API requests per day, we created 10 imgur accounts and divided the task of
We removed those categories that provided no images
and the images with fewer than 10 comments. %  in Imgur and some yield no result with at least 10 comments.
Our eventual dataset consisted of 10K images with 75 categories. % of 80 categories in MSCOCO.
%In terms of data cleaning, we used the same procedure as in section \ref{sec:mscoco}.

%Each node of our graph is a document consisting of the captions belonging to each of mscoco images.
%% Graph edges are the calculated similarity between each two nodes after selecting a cut-off threshold using various trials-and-errors.
%% For measuring the similarities, captions were cleaned using NLP methods from Python NLTK package and Word2Vec[1] implemented in gensim package including
%% removing stopwords, lemmatization using WordNet lemmatizer, removing URLs, non-alphabetical characters, and most common words that would not change the signal strength.
%% Cleaned data results were fed to a model in which embedding vectors were trained on text8 corpus in gensim.
%% Text8 corpus is the first 100MB of flat text words from Wikipedia, has 253855 unique words and 17005208 words in total.
%% In order to calculate the pairwise distance between each two nodes Word Mover’s Distance introduced in 2015 in [2] is used.
%% WMD is an improved edition of Earth Mover’s Distance (EMD) that calculates the minimum travelling distance from one text word embeddings to another.
%% We selected WMD distance because it calculates the semantic distance based on the corpus it has been trained
%% and is more accurate than cosine similarity hence it infers understand Chicago is semantically close to Illinois.
%% WMD also can generalize better than cosine distance for unseen words.

{\bf Setup.} A graph of 10K vertices (i.e., images) with total of 49995k edges was generated %that would take months
by calculating the pairwise similarity between the comments from each image.
%Edge weights were obtained by calculating similarity between each two vertices after selecting a cut-off threshold using various trials-and-errors.
To compute the similarities, the comments were first cleaned (i.e., removing stopwords, URLs and non-alphabetical letters) % and lemmatization)
using %NLP methods from Python NLTK package and Word2Vec[1] implemented in gensim package
natural language toolkit (NLTK) \cite{bird2009natural} and vector embedding using Word2Vec \cite{rehurek_lrec}.
%Cleaned comments were fed to a model where embedding vectors were trained on text8 corpus in gensim.
Then, the sanitized comments were used to compute Word's Mover's Distance (WMD) \cite{kusner2015word}
%which calculates the minimum traveling distance from one text embedding to another. % and yields a semantic distance based on the training corpus.
%and also generalize  than cosine distance for unseen words.
%% We relied on a distributed computing by partitioning our graph generation task into
%% 1250 smaller tasks and merged the end results. % to have a quicker turnaround time.
%We relied on
using HTCondor distributed computing software. % to perform these calculations.
%Later, the edge weights were normalized to the range of $[0~ 1]$.
In our case, the WMD ranged in $(4, 16)$ and we used a Gaussian kernel to transform the WMD into similarity measure within $(0, 1)$. % to construct our graph.
In order to assign object labels in each image,
we used You Only Look Once (YOLO) \cite{redmon2015you},
a deep learning framework %from the open-source deep learning framework named darknet
pretrained on MSCOCO images and categories.
%(using Tesla K40c GPU)
%to detect multiple objects.
After thresholding the confidence level at $40\%$,
we ended up with 6329 images with at least one label. % and no labels on the rest. % had no labels. %were not able to assign any labels on the rest.
%We used Tesla K40c GPU for running the YOLO object detection.

We applied SR framework (using $\alpha = 1$ and $\xi = 0.01$) on this graph with the object labels as measurements on the vertices as in Fig.~\ref{fig:process}.
Note that our framework works in an online manner.
We first select a vertex (i.e., an image) $\pi_1$ and obtain corresponding image labels as in section \ref{sec:selection}
and then run the recovery process as in section \ref{sec:recovery} which will inform how the next vertex $\pi_2$ should be selected.
After running this procedure $m$ times until $\pi_m$, we obtain our policy $\pi$ to be used for final image label recovery.
We will demonstrate our results with $m =50\%$ of the total samples, i.e., selection of $5000$ (of 10K) vertices to obtain image labels and
perform estimation over all vertices including the $5000$ vertices where our model has not obtained a measurement (class/object label).
%We initially began our framework without any image labels,
Note that we do not have ground truth (i.e., true object labels) for this dataset.
We therefore show various interesting qualitative results on the images
where YOLO did not detect objects with high confidence.

{\bf Results.}
Our representative results on object label estimation on the unselected images are demonstrated in Fig.~\ref{fig:result_imgur}.
Note that we were not able to assign any labels for objects in these images using YOLO,
since these objects %in these images
were severely occluded/scaled, not in traditional shape or artificial objects.
%Even for the images that had objects, they had occlusion/scale/illumination issues or they not real objects.
However, our framework successfully suggested labels for some of the unlabeled images with our 75 predefined categories.
%We took out those results where we assign correct labels but the images were either embarrassing or inappropriate.
%For the truely meaningless images, both YOLO and our method did not yield any labels.
For the images where both YOLO and our method did not yield any labels,
post-hoc analysis suggested that many of these images
contained little visual context.
More results are shown in the appendix.

\begin{figure}[!b]
  \vspace{-13pt}
        \centering
        \includegraphics[width = 0.32\columnwidth, height = 0.095\textheight, frame]{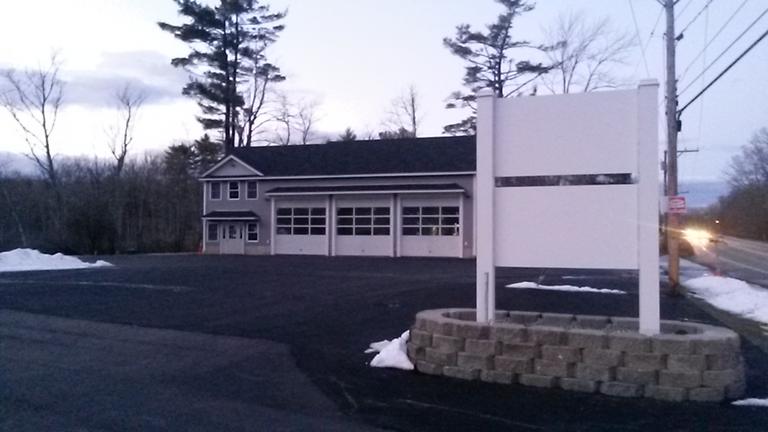}
        \includegraphics[width = 0.32\columnwidth, height = 0.095\textheight, frame]{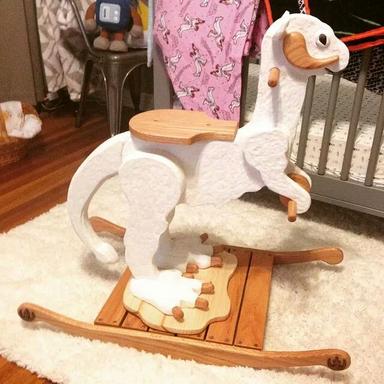}
        \includegraphics[width = 0.32\columnwidth, height = 0.095\textheight, frame]{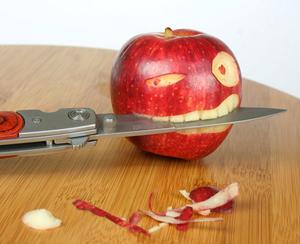}
  \vspace{-9pt}
        \caption{\footnotesize Examples of images where our method assigns false labels.
          We assigned car for body shop (left), sheep instead of sheep shaped chair (middle) and person instead of a person shaped apple (right). }
%  \vspace{-10pt}
        \label{fig:false}
\end{figure}

There were some failure cases where our method assigned false labels that generally falls in one of the following cases:
1) SR predicted ``persons'' but only a small part of a person (e.g., hand, arm or finger) was seen,
2) SR detected objects that had images of texts describing the object,
3) similar/related objects exist in the image but not exact (e.g., car center labeled as `car').  % or a robot labeled as human).
Some of these examples which are still interesting are shown in Fig.~\ref{fig:false}.
%and these examples are given in the supplement.

\vspace{-5pt}
\subsection{Multi-label Learning on MSCOCO Dataset}
\vspace{-5pt}
\label{sec:mscoco}
{\bf Dataset.}
%MSCOCO (Microsoft Common Objects in Context) has 82 categories each of which with more than 5000 labeled instances. MSCOCO dataset has 328000 images with 2500000 labeled instances and is very suitable for 2D localization and learning contextual information compared to other datasets like ImageNet. These images have been collected from Flickr and captioned using Amazon Mechanical Turk.
We used the MSCOCO dataset where $\sim$$328000$ images with $82$ different object categories and relevant captions were available \cite{lin2014microsoft}.
%In order to evaluate our algorithm %and compare the result with other baseline methods
%on a smaller dataset
%Due to computationally challenges,
We retrieved the first 80 images from $80$ different categories and their corresponding captions to generate a smaller dataset to evaluate our SR method.
When overlapping images between categories were discarded, our dataset included $5440$ images.
%% We sampled $\sim$$5000$ images and corresponding captions
%% and created a smaller dataset
%We retrieved the first 80 images from each $80$ different categories, when overlapping images were discarded, we ended up with $5440$ different images.
%The $5440$ images were selected based on ...
%In the end, our datasets includes $5444$ images, captions and object labels in those images sampled from the training dataset in MSCOCO.

%each having 5 lines of caption produced by people.
%Each of these images are also labeled by humans and there are many instances in which multiple objects from MSCOCO categories exist in one image.
%While we tried to include all the categories in the retrieved images, some categories tend to be more popular than the rest,
%e.g., person category with 3045 images and toothbrush with 208 images which still would serve for our purpose in this paper.
% We used pycoco Python API for retrieving and manipulating the data from MSCOCO train2014 annotations json file.
% Considering the fact that the graph consists of 14815846 edges we used distributed computing for reducing the time it takes to get the final result by dividing the job to 1482 smaller jobs.

{\bf Setup.}
A graph using MSCOCO data was generated based on the captions from the $5440$ images (i.e., $5440$ nodes).
%We used MS-COCO captions for images in order to create a similarity graph between the images.
The edges were defined using WMD in the same way as in section \ref{sec:imgur}.
%In order to compute the similarities between vertices, captions were cleaned using Natural Language Toolkit (NLTK) \cite{bird2009natural}
%and Word2Vec implemented in gensim Python packages \cite{rehurek_lrec}.
%including removing stopwords, lemmatization using WordNet lemmatizer and removing swearwords, URL and most common words that would not affect the signal strength.
%The cleaned data were fed to a model in which embedding vectors were trained on text8 corpus in gensim in order to calculate the pairwise distance between each two nodes
%using Word Mover’s Distance (WMD) \cite{kusner2015word} that calculates the minimum traveling distance from one text word embeddings to another.
%WMD is an improved edition of Earth Mover’s Distance (EMD) that calculates the minimum travelling distance from one text word embeddings to another.
%We selected WMD distance because it calculates the semantic distance based on the corpus it has been trained and is more accurate than cosine similarity hence can understand Chicago is semantically close to Illinois.
%WMD also can generalize better than cosine distance for unseen words.
Measurements at each vertex were given as a binary $1 \times 80 $ vector representing object labels
where non-zero elements indicate whether the corresponding objects exist in the image.
Concatenating $5440$ of them, we get a $f_{5440 \times 80}$ matrix which served as the ground truth. %which were regarded as the measurements on the graph vertices.
Depending on the sampling ratio, $m$ number of rows of the matrix were selected according to our policy $\pi$ to obtain object labels,
and we recovered the measurements on all rows. % including the vertices not selected by $\pi$.
%Notice that the ground truth labels are highly skewed, i.e., since there exist only a few objects in each image there the $f$ is very sparse with very few non-zero elements,
%we therefore report precision %area under the curve (AUC) from receiver operating characteristic (ROC) rather than the accuracy to quantitatively evaluate our result.
Notice that the ground truth labels are skewed, i.e., 0s dominates over 1s since there are only a few objects in each image.
Therefore, to evaluate our algorithm, we computed the number of errors that SR makes as well as mean precision of the prediction. % for quantitative evaluation.
We compared our results with two other baseline methods 1) Puy et a.l \cite{puy2015random} and 2) Rao et al. \cite{rao2015collaborative},
which are the state-of-art methods for graph completion.
For the signal recovery step, we used $\alpha = 1$, $\xi = 0.01$ and only $60\%$ of the total bases in our algorithm for estimation
while other methods required all of them.

\begin{figure}[!h]
  \vspace{-8pt}
        \centering
        \includegraphics[width = .85\columnwidth, height = .22\textheight]{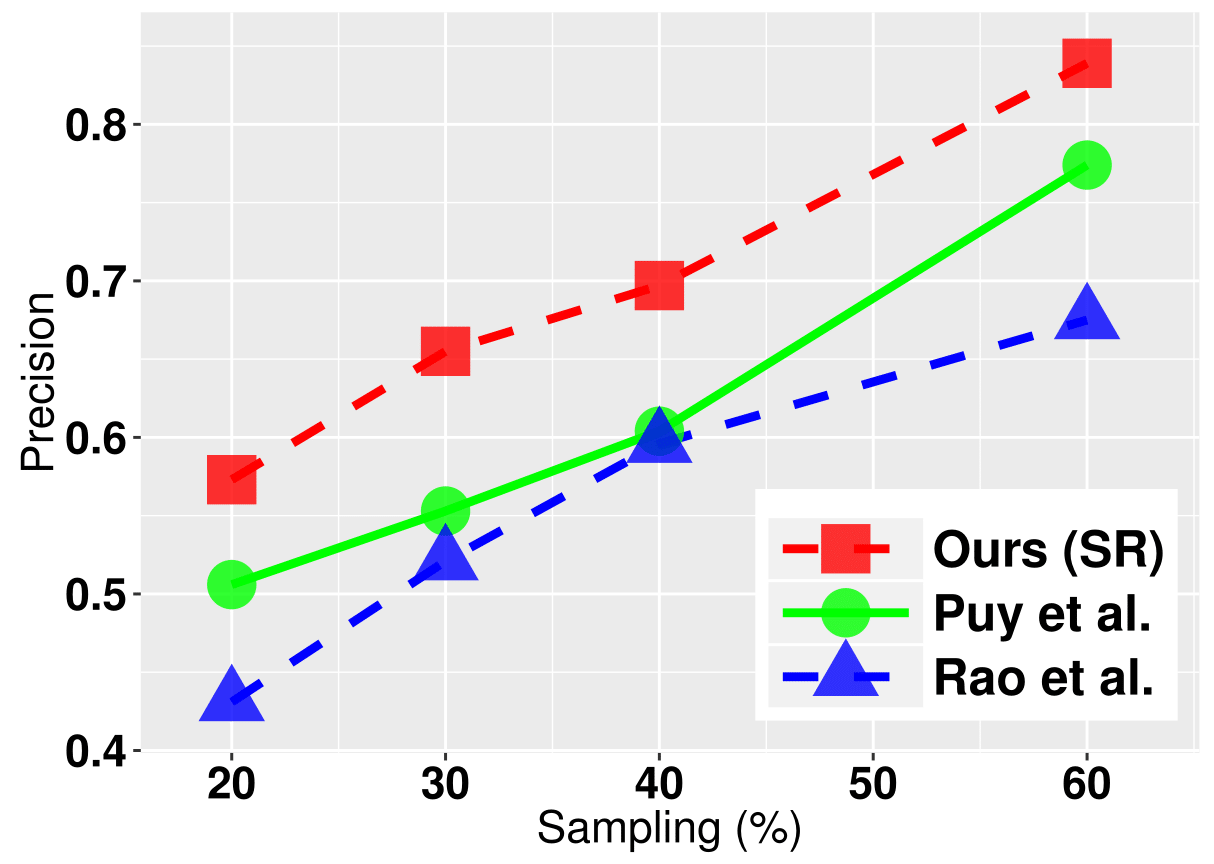}
  \vspace{-10pt}
        \caption{\footnotesize Mean precision over all categories w.r.t sampling ratio.
        As the number of samples increase, precision increases.
        SR (red) shows higher precision than Puy et al (green) and Rao et al (blue) at all sampling rates. }
        \label{fig:precision}
  \vspace{-10pt}
\end{figure}

{\bf Result.}
After recovering the object labels for all images, we thresholded the estimation at $0.15$ to make the recovered labels binary (i.e., 1 if a recovered signal is $>0.15$ and 0 otherwise).
Since baseline methods are stochastic, we ran them 100 times and computed the mean of evaluation scores with optimzed parameters.  %through extensive trials.
Table \ref{tab:error} shows the number of mistakes (out of 435200 estimations) with respect to the size of our policy (or the total number of samples).
As the number of samples that we select increases, the errors decrease in all three methods and our method makes the fewest errors.
\begin{table}[!htb]
  \vspace{-5pt}
\centering
        \small
\scalebox{.95}{
\begin{tabular}{ c| c c c}
  \hline
  Sampling & Ours (SR) & Puy et al. & Rao et al. \\
  \hline
  \hline
  $20\%$ & {\bf 19531} & 21274.6 & 23992.6  \\
  $30\%$ & {\bf 17246} & 19503.3 & 20427.7 \\
  $40\%$ & {\bf 15003} & 17862.2 & 17762.4 \\
  $60\%$ & {\bf 8992} & 10689.6 & 11906.9  \\
  \hline
\end{tabular}}
  \vspace{-8pt}
 \caption{\footnotesize $\#$ of errors (out of 435200) in the recovered measurements %w.r.t. sampling ratio.}
}
  \label{tab:error}
  \vspace{-10pt}
\end{table}
We also report mean precision over all categories instead of accuracy in Fig.~\ref{fig:precision}.
%Although the number of mistakes that we make was the smallest at $25\%$, we can see that is precision is the worst of all.
Here, the precision increases as the size of the policy increases and
our result shows higher precision than those from the baselines as well as in \cite{kim2016adaptive}
reaching up to $0.84$ with $60\%$ of total vertices.

%% when thresholded at 0.15,

%% % 20%
%% Puy has on average of 21274.6 false labels out of 435200.
%% Rao has on average of 23992.6 false labels out of 435200.
%% Ours have 19531 out of 435200.

%% % 30%
%% Puy has on average of 19503.3 false labels out of 435200.
%% Rao has on average of 20427.7 false labels out of 435200.
%% Ours have 17246 out of 435200.

%% % 40%
%% Puy has on average of 17862.15 false labels out of 435200.
%% Rao has on average of 17762.4 false labels out of 435200.
%% Ours have 15003 out of 435200.

%% % 60%
%% Puy has on average of 10689.6 false labels out of 435200.
%% Rao has on average of 11906.9 false labels out of 435200.
%% Ours have 8992 out of 435200.

%% precision 60 40 30 20 \\
%% Ours : 0.839 0.697 0.655  0.237\\
%% Puy : 0.774 0.604 0.553  0.506\\
%% Rao : 0.675 0.596 0.521  0.431\\

\vspace{-5pt}
\subsection{Estimation of PIB Measures using CSF}
\vspace{-5pt}
%% Measures from PIB-PET scans
%% and CSF measures are known to be highly (negatively) correlated \cite{fagan2006inverse}.
%by constructing a graph using CSF measures and applying our framework.

{\bf Dataset.} Our AD dataset includes 79 participants where both PIB-PET scans and CSF are available.
The voxel intensities of PIB-PET scans measure amyloid plaque pathology in the brain
which is highly related to brain function as do the CSF measures,
and these two measures are known to be highly (negatively) correlated \cite{fagan2006inverse}.
We parcellated the brain into multiple regions of interests (ROI) and took the mean of
the PIB measures in 16 selected ROIs to obtain ROI specific PIB measures.
From the CSF data, we obtained various protein levels for each participant.
More details of the dataset are given in the appendix.
%% Although the PIB measures are derived from brain and CSF measures are obtained from cerebrospine,
%% these two measure are known to be highly (negatively) correlated, and CSF measures are often
%% acquired as a surrogate for PET scans due to its cost.
%% We therefore construct a graph using the CSF measure, select a subset of the participants
%% to acquire PIB measure, then estimate the PIB measure over all cohort.

{\bf Setup.}
The PIB images and the CSF measures involve different costs where PET scans are much more expensive,
and CSF measures are often acquired as a surrogate for PET scans.
In this experiment, we try to estimate PET image-derived measures based on CSF measures from the full cohort and PET image-derived measures on a subset of participants.
A graph using CSF measures from each participant (i.e., vertex) was created by measuring similarity (i.e., edge) between participants using a Gaussian kernel $\exp(-(x-y)^2/\sigma^2)$ with $\sigma = 1$.
Then we applied our framework as in Alg. \ref{alg:alg1} %to adaptively decide a policy $\pi$, obtain PIB measures on the 16 ROIs (i.e., 16 features) from $v \in \pi$ and recover the PIB measures over all vertices $v \in V$ (i.e., all participants).
to decide a policy to obtain PIB imaging measures from $v \in \pi$ on the 16 ROIs and recover the measures over all (remaining) participants.
We used $\xi$$=$$0.01$ for the sparsity parameter, $k$$=$$50$ number of eigenvectors and $\alpha$$=$$1$ for the signal recovery step.

%% \begin{figure}[!t]
%%   \centering
%%   \includegraphics[width = .47\textwidth]{fig/est_freq_30}
%% %  \includegraphics[width = .23\textwidth]{fig/est_freq_50}
%%   \caption{Estimation of the graph Fourier coefficients with $30\%$ of the total samples. The error between the ground truth (black circle) and our estimation (blue) is smaller than the estimations using Puy et al (red) and Rao et al (green).}
%% \label{fig:roi_error}
%% \end{figure}

\begin{figure}[!htb]
\vspace{-8pt}
  \centering
  \includegraphics[width = \columnwidth]{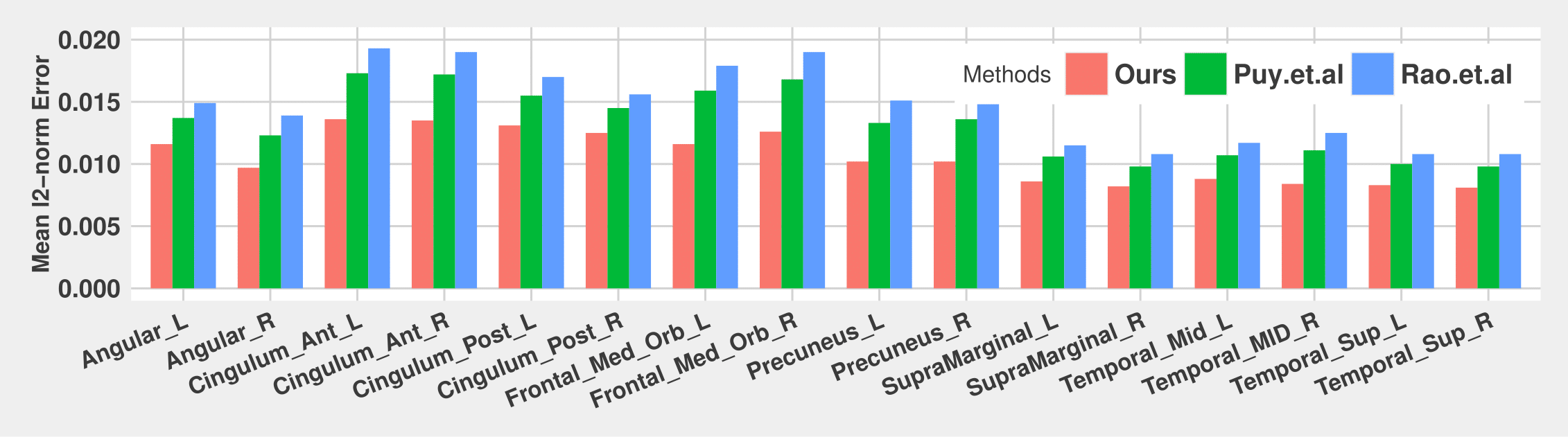} %roi_error_30}%  \hspace{2mm}

  \includegraphics[width = \columnwidth]{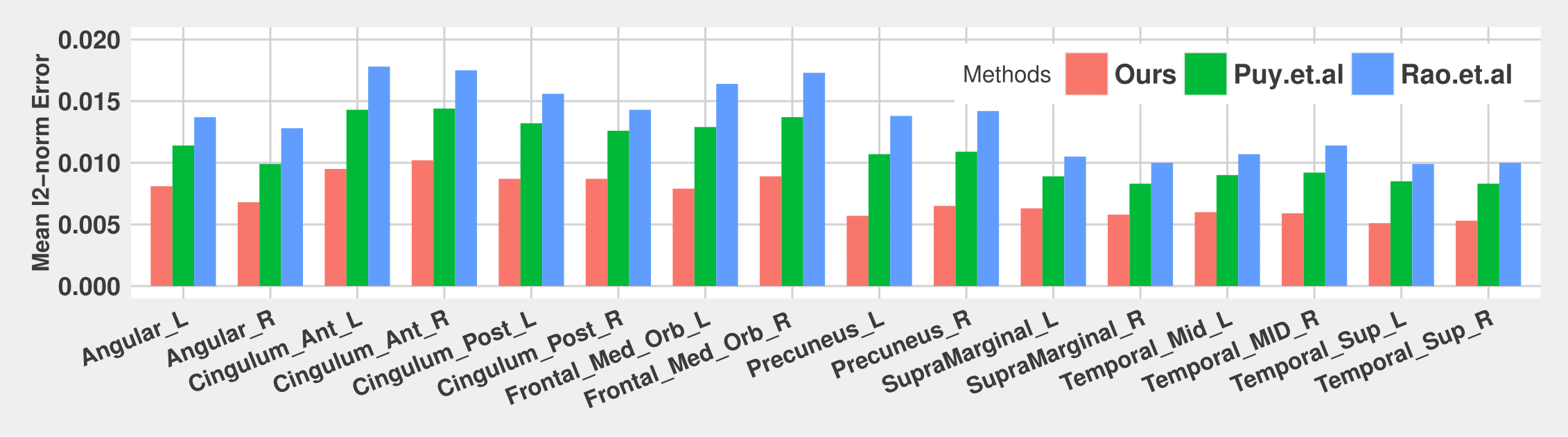} %roi_error_30}

\vspace{-8pt}
  \caption{\footnotesize ROI-wise mean $\ell_2-$norm error between recovered signals and the ground truth using SR (red), Puy et al.(green) and Rao et al.(blue).
    Top: using $30\%$ of the total samples,
    Bottom: using $50\%$ of the total samples.
%    SR shows the lowest at all ROIs.}
}
\label{fig:roi_error}
\vspace{-10pt}
\end{figure}

{\bf Result. }
We show the $\ell_2$-norm of the error between the ground truth and the recovered measures for evaluation.  %and compared with the results from other baseline methods.
Again, we ran the baseline methods $500$ times to compute the mean of errors due to their stochasticity.
We ran the experiments by varying $m$ and reported the results with $m = \{30\%, 50\%\}$ of the total samples.
As summarized in Fig.~\ref{fig:roi_error}, our result (in red) shows much lower error than the baseline methods.
When we used these estimation results to identify whether each participants had elevated amyloid burden
(i.e., whether mean of PIB measures over all ROIs is $>1.18$),
our estimation offered $91.1\%$ accuracy while \cite{puy2015random} and \cite{rao2015collaborative} provided $88.6\%$ and $87.6\%$. %, slightly lower than ours.
%When $30\%$ of the total samples were used, the mean $\ell2-$norm errors in the frequency space between the ground truth and the estimations using our method, Puy et al and Rao et al were $1.45$, $3.54$ and $3.84$ respectively.

\vspace{-8pt}
\section{Conclusion}
\vspace{-8pt}

Motivated by various instances in modern computer vision that
involve an interplay between the data (or supervision) acquisition
and the underlying inference methods, we study the
problem of adaptive completion of a multivariate signal obtained sequentially, on the vertices of
graph.
%large scale data acquisition process that requires human supervision,
%we proposed a method that adaptively queries samples to acquire data from (represented as graph vertices)
%in a sequential manner
%by interacting with a multi-variate signal recovery algorithm on graphs.
By expressing the optimization in the frequency domain of the graph, we show
how a simple algorithm based on adaptive submodularity yields
impressive results across diverse applications.
%Using recent work on adaptive submodularity,
%we showed how our problem can be formulated using diminishing returns
%and solved via an efficient algorithm utilizing the graph Fourier/wavelet transform.
%We provided various experimental results using our method o
On large-scale vision datasets,
our proposal complements object detection algorithms %the ability of object detectors and categorization algorithms,
by solving a completion problem %sequentially
(using auxiliary information).
The model provides promising evidence how neuroimaging studies under %significant
budget constraints can be conducted (in a sequential manner) with minimal deterioration in statistical power.
%to predict object labels in the images, which demonstrated interesting results by successfully predicting
%the labels even when a state-of-art object detector was not able to detect objects.
%We've also shown that our method can be adopted for experimental design in neuroimaging applications
%which would help save substantial cost in the data acquisition step
%by guiding which participants to obtain measurements and accurately estimating the measurements on the participants
%who has not been selected.
Our open source distribution will enable applications to other 
settings in vision which involves partial measurements and/or sequential observations of
data structured as a graph.

\vspace{-8pt}
\section{Acknowledgement}
\vspace{-8pt}
This research was supported by NIH grants AG040396, AG021155, EB022883, NSF CAREER award 1252725, UW
ADRC AG033514, UW CIBM 5T15LM007359-14, and UW CPCP AI117924.

{\small
\bibliographystyle{ieee}
\bibliography{wonhwa_ref}
}

\end{document}